\pdfoutput=1

\documentclass[11pt]{article}

\usepackage[]{EACL2023}

\usepackage{times}
\usepackage{latexsym}

\usepackage[T1]{fontenc}

\usepackage[utf8]{inputenc}

\usepackage{microtype}

\usepackage{inconsolata}

\usepackage{booktabs}
\usepackage{graphicx}
\usepackage{url}

\title{Large Scale Multi-Lingual Multi-Modal Summarization Dataset}

\author{Yash Verma* \\
  IISER Kolkata\\
  \texttt{yashv7523@gmail.com} \\\And
  Anubhav Jangra* \\
  Indian Institute of Technology Patna \\
  \texttt{anubhav0603@gmail.com} \\\AND
  Raghvendra Kumar\\
  Indian Institute of Technology Patna \\
  \texttt{raghvendra.kumar1004@gmail.com}\\\And
  Sriparna Saha\\
  Indian Institute of Technology Patna\\
  \texttt{sriparna@iitp.ac.in}}

\begin{document}
\maketitle
\def\thefootnote{*}\footnotetext{These authors contributed equally to this work.}\def\thefootnote{\arabic{footnote}}

\begin{abstract}
Significant developments in techniques such as encoder-decoder models have enabled us to represent information comprising multiple modalities. This information can further enhance many downstream tasks in the field of information retrieval and natural language processing; however, improvements in multi-modal techniques and their performance evaluation require large-scale multi-modal data which offers sufficient diversity. Multi-lingual modeling for a variety of tasks like multi-modal summarization, text generation, and translation leverages information derived from high-quality multi-lingual annotated data. In this work, we present the current largest multi-lingual multi-modal summarization dataset (\texttt{M3LS}), and it consists of over a million instances of document-image pairs along with a professionally annotated multi-modal summary for each pair. It is derived from news articles published by \texttt{British Broadcasting Corporation(BBC)} over a decade and spans 20 languages, targeting diversity across five language roots, it is also the largest summarization dataset for 13 languages and consists of cross-lingual summarization data for 2 languages. We formally define the multi-lingual multi-modal summarization task utilizing our dataset and report baseline scores from various state-of-the-art summarization techniques in a multi-lingual setting. We also compare it with many similar datasets to analyze the uniqueness and difficulty of \texttt{M3LS}. \footnote{The dataset and code used in this work are made available at \url{https://github.com/anubhav-jangra/M3LS}.}
\end{abstract}

\section{Introduction}
The world we live in today is very diverse, with over 7,000+ languages spoken across the globe\footnote{\url{https://www.ethnologue.com/guides/how-many-languages}}. These languages have varying traits and are spoken by communities of various sizes depending upon the popularity of the language. For example, Mandarin consists of over 50,000 \textit{hanzi} (characters) and is spoken by over 1.117 billion people\footnote{\url{https://www.berlitz.com/en-uy/blog/most-spoken-languages-world}}, while there exist languages like Rotokas, which is an indigenous language spoken by about 4,320 people on the island of Bougainville, Papua New Guinea, which consists of only 12 letters\footnote{\url{https://en.wikipedia.org/wiki/Rotokas_language}}.

\begin{table}[]
\tiny
\centering
\caption{Comparison of proposed dataset with existing large-scale multi-lingual and multi-modal datasets.}
\label{tab:comparison}
\vspace{0.7em}
\resizebox{7.8cm}{!}{
\begin{tabular}{cccc}
\toprule
\multicolumn{4}{c}{\textbf{Multi-lingual Summarization Datasets}} \\
\hline
\multicolumn{1}{c}{\textbf{Dataset Name}} & \multicolumn{1}{c}{\textbf{Dataset Size}} & \multicolumn{1}{c}{\textbf{\#Languages}} & \multicolumn{1}{c}{\textbf{Domain}} \\ 
\hline
\texttt{XL-Sum} \cite{hasan2021xl} & 1M & 44 &  News \\
\texttt{MLSUM} \cite{scialom2020mlsum} & 1.5M  & 5 & News \\
\texttt{WikiLingua} \cite{ladhak2020wikilingua} & 770K & 18 & Tutorials \\
\texttt{MLGSum} \cite{wang2021contrastive} & 1.1M & 12 & News \\
\hline
\texttt{M3LS} (Ours) & 1.1M & 20 & News \\
\hline
\multicolumn{4}{c}{\textbf{Multi-modal Summarization Datasets}} \\
\hline
\multicolumn{1}{c}{\textbf{Dataset Name}} & \multicolumn{1}{c}{\textbf{Dataset Size}} & \multicolumn{1}{c}{\textbf{Modalities}} & \multicolumn{1}{c}{\textbf{Domain}} \\ 
\hline
\texttt{MSMO} \cite{zhu2018msmo} & 314K & Text + Image & News \\
\texttt{E-DailyMail} \cite{chen2018abstractive} & 219K &  Text + Image & News\\
\texttt{How2} \cite{sanabria2018how2} & 190K & Text + Video + Audio & Multiple Domains \\
\texttt{MMSS} \cite{li2018multi} & 66K &  Text + Image & News \\
\texttt{VMSMO} \cite{li2020vmsmo} & 185K &  Text + Video + Audio  & Social Network\\
\hline
\texttt{M3LS} (Ours) & 1.1M & Text + Image & News\\
\bottomrule
\end{tabular}
}
\end{table}

These languages, although very crucial, restrict people to communicate their thoughts to others who speak the same language. The gift of sight, however, is something that is universally shared by every human being on this plant, irrespective of their culture, ethnicity, or the language that they speak. Through this work we aim to instigate the research towards improving existing automatic summarization systems by leveraging information from multiple languages and visual modalities. 

Various studies in the past have illustrated how unified summarization frameworks across multiple languages improve the summarization quality over mono-lingual frameworks \cite{wang2021contrastive}. Similarly, there have been works in multi-modal summarization that illustrate how multi-modal input can help improve the quality of summarization over text summarization systems \cite{jangra2020text,jangra2020multi,chen2018abstractive,mukherjee-etal-2022-topic}. Additionally, having multiple modalities in the output summary can help improve the overall satisfaction of the user \cite{zhu2018msmo,jangra2021multi}. Multiple modalities can also compensate for the inability of individual modalities to express various aspects of the summary. For instance, it is hard to express abstract concepts like ``freedom", ``gravity", etc. through images, while it can be expressed through text conveniently. Similarly, it is very difficult to describe a ``Pangolin" to someone who hasn't seen one beforehand. 

Hence, in this work we propose the task of Multi-modal Multi-lingual Summarization (\texttt{M3LS}), and also release the \texttt{M3LS} dataset\footnote{A sample of our dataset is available at \url{https://github.com/zenquiorra/M3LS}, the complete dataset will be released in the camera ready version of the work} to facilitate the research in this direction. The dataset comprises 1.1M news articles, spanning 20 languages comprising \textit{English}, \textit{Chinese}, \textit{Spanish}, \textit{Russian}, \textit{French}, \textit{Ukrainian}, \textit{Portuguese}, \textit{Japanese}, \textit{Tamil}, \textit{Hindi}, \textit{Marathi}, \textit{Gujarati}, \textit{Bengali}, \textit{Sinhala}, \textit{Urdu}, \textit{Pashto}, \textit{Indonesian}, \textit{Telugu}, \textit{Punjabi}, and \textit{Nepali}; making it the largest language-spanning summarization dataset. To the best of our knowledge, the proposed dataset is the largest summarization dataset for 13 languages (\textit{Russian}, \textit{Ukrainian}, \textit{Tamil}, \textit{Hindi}, \textit{Marathi}, \textit{Gujarati}, \textit{Bengali}, \textit{Sinhala}, \textit{Urdu}, \textit{Pashto}, \textit{Telugu}, \textit{Punjabi}, and \textit{Nepali}).

We hope that the proposed task and the dataset will instigate and inspire multi-modal and multi-lingual research in less-explored languages for solving various tasks including but not limited to automatic summarization \cite{Nallapati2016AbstractiveTS,DBLP:journals/corr/SeeLM17}, article headline generation \cite{jin2020hooks,gavrilov2019self,zhang2018question}, keyword extraction \cite{showrov2019keyword,lee2008news,yao2019research}, image caption generation \cite{xu2015show,bai2018survey}, multi-modal embedding generation \cite{sun2019videobert,lu2019vilbert,li2019visualbert,zhou2020unified}, large-scale language modeling \cite{raffel2020exploring,devlin2018bert} etc.

The major contributions of this work are as follows - \textit{1) We have proposed the multi-modal multi-lingual summarization (\texttt{M3LS}) task. 2) We have released the largest multi-modal summarization dataset that spans 20 languages. 3) The proposed dataset is the largest text summarization dataset for 13 languages. 4) To the best of our knowledge, we present the first ever multi-modal cross-lingual dataset (consisting of Japanese-to-English and English-to-Japanese). 5) We have provided multi-modal summarization baseline results for our dataset and a detailed analysis of the dataset.}

\section{Related Work}

The field of text summarization is more than 5 decades old \cite{Edmundson1969NewMI}, and has evolved to a great extent in recent years. Prior to the advances in sequence-to-sequence frameworks \cite{sutskever2014sequence}, people mainly focused on extractive summarization techniques that aim to generate summary via extracting words, phrases, or sentences \cite{mihalcea2004textrank, saini2019extractive, alguliev2010multi}. 
\citet{DBLP:journals/corr/SeeLM17} proposed the Pointer-Generator Networks, an attentive recurrent neural network based framework \cite{bahdanau2015neural}. 
Recent years have seen great progress in research in automatic summarization leveraging transformer based models \cite{zhang2020pegasus,devlin2018bert} and attention mechanism \cite{vaswani2017attention}. 

In this section we discuss the related works showcasing multi-modal datasets and multi-lingual datasets. 
A detailed size comparison of these datasets with \texttt{M3LS} is shown in Table \ref{tab:comparison}.



\subsection{Multi-modal summarization datasets}

Multi-modal summarization is the task of summarizing content comprising two or more input modalities. The output can be uni-modal or multi-modal depending on the task. In this section, we discuss existing large-scale multi-modal summarization datasets proposed in the community. We point the readers to \citet{jangra2021survey} for a comprehensive survey.

\textbf{MSMO}: \citet{zhu2018msmo} proposed a multi-modal summarization dataset that consists of text and images. The dataset is obtained from the \texttt{DailyMail}\footnote{\url{https://www.dailymail.co.uk/home/index.html}} website and contains 314,581 instances in English language. 
However, \citet{hasan2021xl} illustrated that the \texttt{DailyMail} news highlights lack novel n-grams. \citet{fabbri2021summeval} also highlighted the inconsistency in quality of some reference summaries in the \texttt{CNN/DailyMail} dataset \cite{Nallapati2016AbstractiveTS}. 

\textbf{E-Dailymail} \citet{chen2018abstractive} proposed the E-Dailymail dataset, which contains text and images extracted from the \texttt{DailyMail} website. The dataset consists of 219,100 instances in English, containing the input document, article title, images, and image captions. 

\textbf{How2}: \citet{sanabria2018how2} proposed a multi-modal summarization dataset consisting of text, video, and audio modalities; it contains over 2000 hours of videos accompanied by the corresponding audio and speech transcriptions. 


\textbf{MMSS}: \citet{li2018multi} proposed a multi-modal summarization dataset consisting of text and images with the aim of proposing an image-aided sentence summarization framework. 
The dataset has 66K instances in English language, that is generated by extracting sentence-headline pairs from the \texttt{Gigaword} corpus\footnote{\url{ https://github.com/harvardnlp/sent-summary}}. 


\textbf{VMSMO}: To the best of our knowledge, \citet{li2020vmsmo} proposed the first large-scale asynchronous text-audio-video summarization dataset. The dataset is generated from the famous microblogging platform \texttt{Sina Weibo}\footnote{http://ir.weibo.com/}, and comprises of 184,920 instances in Chinese language. 

Similar trends of incorporating multiple modalities in language tasks can also be noticed in several tasks like question answering \cite{singh2021mimoqa}, translation \cite{elliott2017imagination}, sentiment analysis \cite{soleymani2017survey}, lexico-semantic classification \cite{jha2022combining}, keyword extraction \cite{verma2022maked} etc.


\subsection{Multi-lingual Text Summarization Datasets}

The popularity of studying the benefits of summarization in different languages to improve summarization qualities increased over the past few years. There have been a lot of research work in bi-lingual setting; however, in this work, we limit ourselves to discussing multi-lingual summarization datasets to be concise. 



\textbf{MLSUM} : \citet{scialom2020mlsum} proposed the \texttt{MLSUM} dataset that consists of 1.5 million news articles obtained from the \texttt{Dailymail/CNN} websites. The dataset spans five languages - French, German, Spanish, Russian and Turkish. 


\textbf{XL-Sum}: \citet{hasan2021xl} proposed the \texttt{XL-Sum} dataset that consists of 1.35 million articles in 44 languages obtained from \texttt{BBC news}, making it the most language-diverse summarization dataset to date. However, 25 of these 44 languages do not contain even 10,000 instances, making it incompetent to train any language model. 


\textbf{WikiLingua}: \citet{ladhak2020wikilingua} proposed the \texttt{Wikilingua} dataset, which is the largest parallel multi-lingual summarization to date. The dataset consists of 770K instances in English language, and is extended to 17 other languages for varying number of English articles. 


\textbf{MLGSum}: \citet{wang2021contrastive} proposed the \texttt{MLGSum} dataset that consists of articles from various news providers such as \texttt{BBC}, \texttt{france243} and \texttt{select faz}. The dataset has five high-resource and seven low-resource languages, with a total of 1.1 million instances, and is a rich source for text summarization for German language with ~500K instances. 

We observe that multiple popular datasets (see Table \ref{tab:comparison}) in multi\-modal summarization and multi-lingual summarization are useful for both technique evaluation and technique improvisation. However, the combined field of multi\-lingual multi\-modal summarization has remained largely unexplored, and it can be attributed to the lack of dedicated high quality dataset and formalizing it as a problem statement. Hence, we formally define the \texttt{M3LS} task and discuss the dataset addressing the problem further.




\section{M3LS Task}
Given for each language $l_k \in L$ where $L$ is the set of all languages, we have data $M^{l_k} = <T^{l_k} , I^{l_k}>$, where $T^{l_{k}}=\left\{t_{1}^{l_{k}}, t_{2}^{l_{k}}, \ldots, t_{|T|}^{l_{k}}\right\}$ is a set of documents, and $I^{l_k}=\{ I^{t_{1}^{l_{k}}}, I^{t_{2}^{l_{k}}}, \ldots, I^{t_{|T|}^{l_{k}}}\}$ is a set of images, where $I^{t_{j}^{l_{k}}}=\left\{i_{1}, i_{2}, \ldots, i_{|I|}\right\}^{t_{j}^{l_{k}}}$ denotes the set of images belonging to the document $t_j^{l_k} \in T^{k_k}$ and $|.|$ denotes the cardinality of a set.

The task is to obtain a function $F$ that maps documents $t_{j}^{l_{k_1}} \in T^{l_{k_1}}$ in language $l_{k_1}$ along with their corresponding images, $I^{t_j^{l_{k_1}}} \in I^{l_{k_1}}$ to a set of multi-modal summaries in target language, $l_{k_2}$, comprising of text summaries (denoted by $O^{l_{k_2}}$) along with images from the input (denoted by $I^{l_{k_1}}$).
\begin{equation}
    F:<T^{l_{k_1}}, I^{l_{k_1}}> \rightarrow  <O^{l_{k_2}}, I^{l_{k_1}}>
\end{equation}

\begin{figure*}[]
    \centering
    \includegraphics[width=0.7\textwidth]{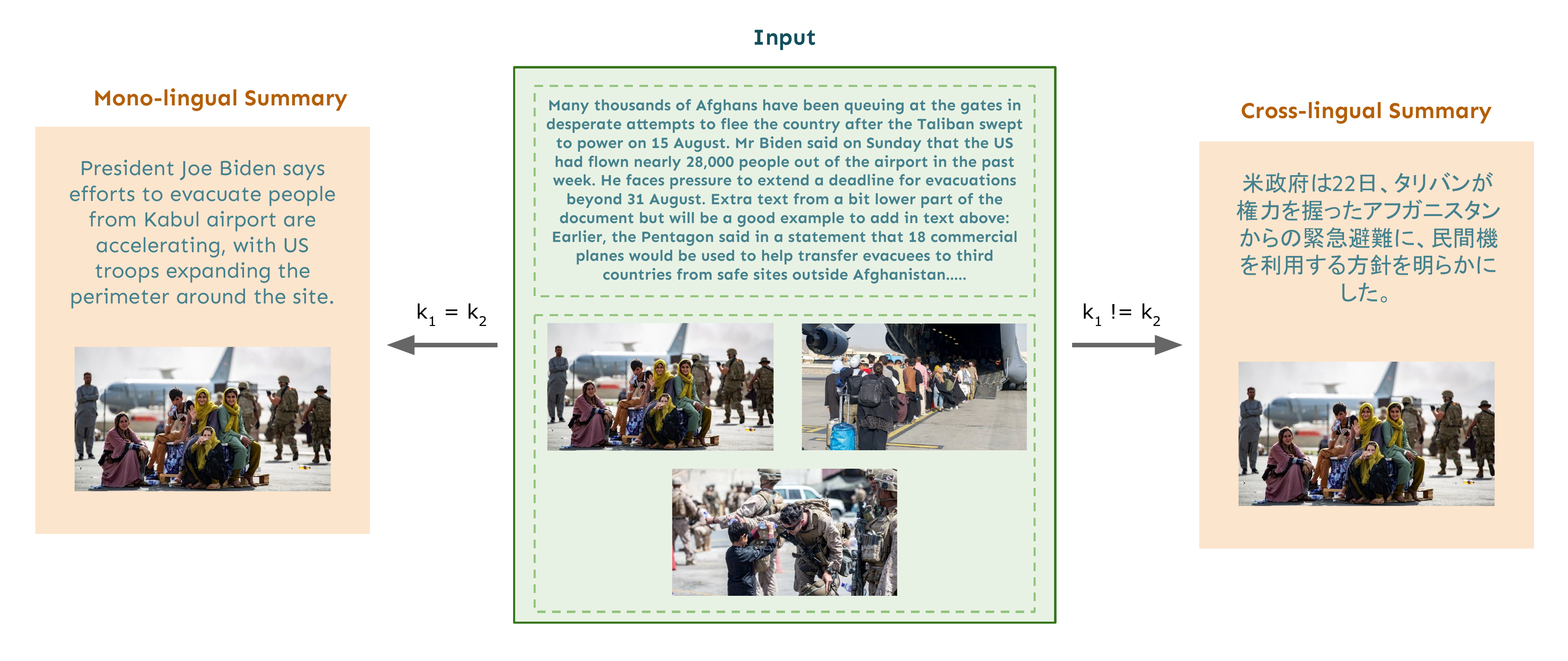}
    \vspace{-4mm}
    \caption{Proposed \texttt{M3LS} task.}
    \label{fig:box}
    \vspace{-1em}
\end{figure*}

When $k_1\neq k_2$, we have multi-modal cross-lingual summarization, otherwise the task is multi-modal mono-lingual summarization, a graphic representation of the task is shown in Figure \ref{fig:box}.

\section{M3LS Dataset}

Through the \texttt{M3LS} task, we motivate the need for a multi-modal multi-lingual dataset by studying the developments in summarization techniques such as secondary enhancements using images with multi-modal output \cite{zhu2018msmo}, video-based multi\-modal summarization \cite{li2020vmsmo} and using multi\-objective optimization \cite{jangra2020multi}. On the other hand, development of multi-lingual transformer based models like \citet{xue2020mt5} has publicly available checkpoints fine-tuned for multiple language modelling tasks, including multi\-lingual summarization.

Development of such models requires high-quality heterogeneous data and improvements in various models utilizing multi-modal shared attention layers for annotated data with image-text pairs  for a specific language task. To address these issues, we present \texttt{M3LS} and in this section we discuss various steps involved in its construction.

\begin{figure}[h]
    \includegraphics[width=0.4\textwidth]{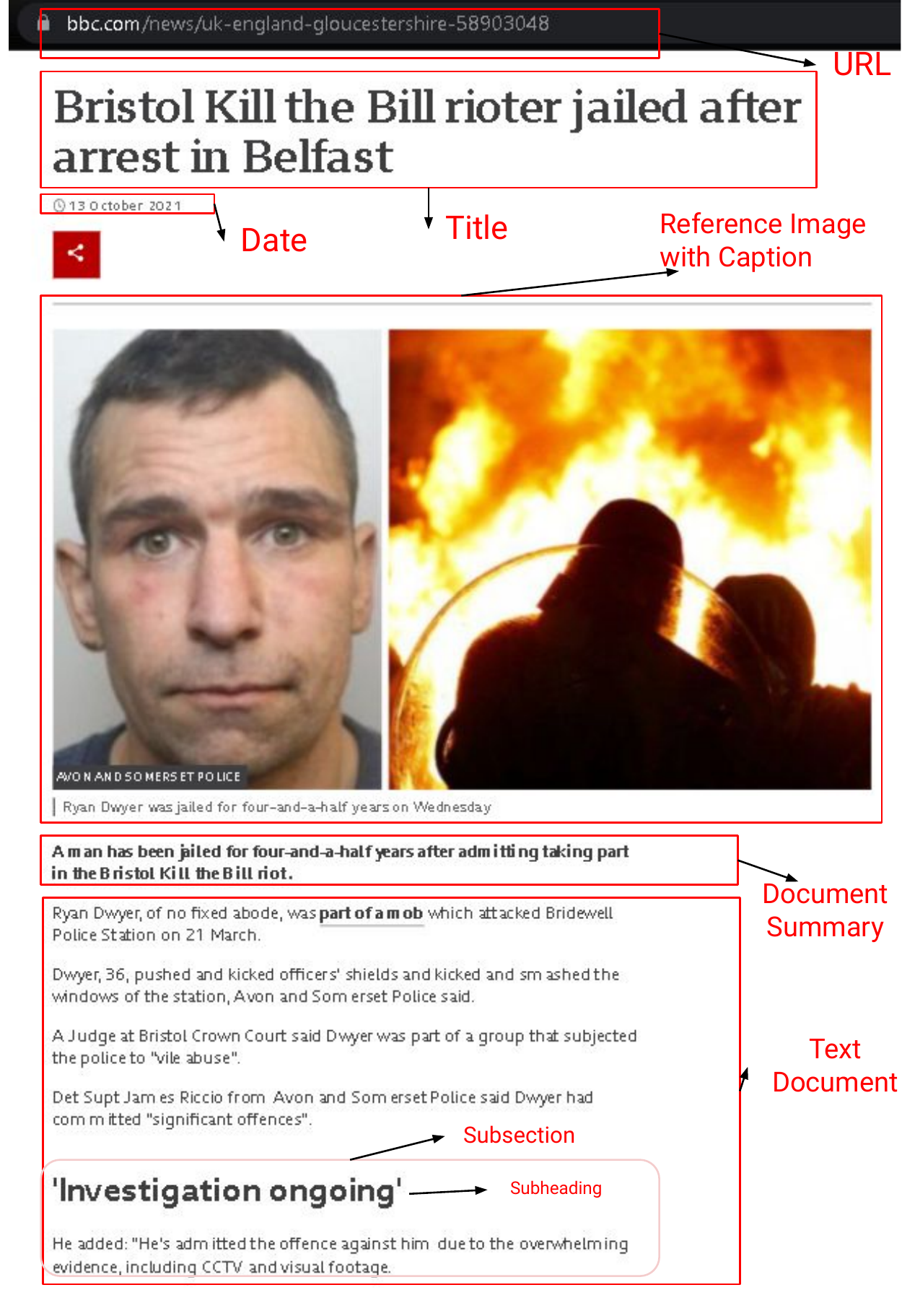}
    \caption{Snapshot format of a webpage used in development of M3LS, and various features extracted during the scraping procedure}
    \label{fig:webpage}
\end{figure}

\subsection{Dataset Construction}

We explore the news domain, as it is one of the most abundant and readily available domains and covers articles in multiple topics, while describing the events and lacking extreme bias. We analyzed the structure of articles and surveyed multiple news providers before finalizing on \texttt{BBC News}, which provides full sentence summaries in a uniform structured format across multiple languages. The summaries are professionally created by the author which ensures the quality of the data. We explain the steps involved in creating the \texttt{M3LS} dataset and discuss various aspects of the data.

{\bf BBC News}: \texttt{BBC News}\footnote{\url{https://www.bbc.com/news}} is a division of British Broadcasting Corporation responsible for gathering and broadcasting current news affairs. Each BBC news article has a text summary comprising complete sentences in the present tense, avoiding opinions and sensationalism. We cover 20 different languages with summaries written in corresponding languages. We extract data from various parts of the webpage as shown in Figure \ref{fig:webpage}.

{\bf Obtaining Articles}: We obtain links to articles from the corresponding \texttt{Twitter\footnote{\url{https://twitter.com/bbc}}} pages for each BBC language news dataset.  
To extend the dataset, we scrape\footnote{Data is collected in accordance with the terms and conditions mentioned on the website} valid links\footnote{A link is valid if it contains a BBC article summary for the corresponding domain.}  obtained from the parsed articles of each language.

The final collection of links is scraped separately using \texttt{scrapy}\footnote{\url{https://scrapy.org}} to obtain the final dataset. Since these links are showcased at the corresponding \texttt{twitter} page, these links ensure articles containing topics of interest and high popularity. We extend the dataset by recursively extracting links from suggestions and hyperlinks within a webpage.

\textbf{Structuring the Data}: We obtain various features from the webpage as shown in Figure (\ref{fig:webpage}), and compile them in a \texttt{JSON} format; we also provide a dedicated parser, instructions and a tutorial for the ease of access of features from any instance. The data is freely available for use in accordance with the terms and conditions of \texttt{BBC News}, we discuss this in detail on the same link where our dataset is uploaded. 

{\bf Text Validation}: In order to ensure high-quality text from the source, we manually read 10 instances from each language\footnote{For languages unknown to the authors, we use \texttt{Google Translate}\url{https://translate.google.com} to translate the content in English language} from the collected links  to verify if the articles are descriptive in nature and consist of text written in complete sentences. 


\noindent\textsc{\bf Summary Validation}: 
We manually checked the summary quality for 100 articles each in 4 languages\footnote{We restrict ourselves to 4 languages (\texttt{English}, \texttt{Hindi}, \texttt{Bengali} and \texttt{Marathi}) due to the understanding of languages of the authors presenting this work} from our dataset and validated if the given summary captures the information represented in the text. For every article, after carefully reading the text, we assign the gold summary a score between 1-5 with 5 being the best possible summary which captures most of the important information from the given article and vice versa, and also including parameters like summary length and length of the article. We observe that for > 70 articles across the languages evaluated obtain a score of > 4 out of 5 in our analysis. 
Assuming the uniformity of articles published by BBC across multiple domains, we assume that this fact is true for every language in our dataset.

\noindent\textsc{\bf Final Dataset}: In final dataset, each news article contains the text document, images with corresponding captions, keywords, links to related news articles, and a multi-modal summary comprising of a few sentences and an image.

\noindent\textsc{\bf Cross-lingual Dataset}: Our cross-lingual dataset contains all features from our final dataset, along with multi-modal summaries consisting of text in another language. It is obtained from the links given by the author within the Japanese language article to the corresponding article in English language, we manually check the information provided in both articles using \texttt{Google Translate}\footnote{\url{https://google.com/translate}} for 100 instances to verify the similarity of the content and summaries provided.


\noindent\textbf{Train-Test-Validation split:} The dataset has 1.2 million news articles which we split into 80\% training, 10\% test and 10\% validation for languages having $\leq$ 50,000 articles, otherwise we select 90\% data for training, 5\% for testing and 5\% as validation split. 






\section{Dataset Analysis}

\subsection{Overview}
The \texttt{M3LS} dataset has 1.11M+ multi-lingual multi-modal instances across 20 languages and over 9K cross-lingual multi-modal instances for \texttt{English}-\texttt{Japanese} language pair. The dataset can be categorized into 8 high resource languages and 12 low resource languages\footnote{The categorization is done based on a threshold value of 50k data instances.} (refer to Appendix \ref{sec:appendix-B} for more details). The chosen languages originate from different parts of the globe, and belong to 5 different language roots: \textit{Indo-European}, \textit{Austronesian}, \textit{Japanic}, \textit{Dravidian}, and \textit{Sino-Tibetan}.

\texttt{M3LS} dataset is quite diverse, with the least \#articles for \texttt{Sinhala} (10,148) and greatest \#articles for \texttt{English} (376,367). The dataset becomes even more complex and challenging with different sizes of input documents for different languages, with document size varying from \~ 330 tokens to over 2800+ tokens. The dataset articles cover a wide time span, with articles from 2009 to 2021 (refer to Appendix \ref{sec:appendix-A} for more details).

We hope that the \texttt{M3LS} dataset will instigate and inspire research in less-explored languages, since 14 out of these 20 languages covered by the dataset are among the top-20 most spoken languages in the world\footnote{\url{https://lingua.edu/the\%2D20\%2Dmost\%2Dspoken\%2Dlanguages\%2Din\%2Dthe\%2Dworld\%2Din\%2D2022/}}; this diversity helps in modelling tasks for both well-explored and less-explored languages.

\subsection{Dataset Comparison}
To study the size and span of our dataset, we compare \texttt{M3LS} with other summarization datasets extracted from the \texttt{BBC News} domain. We found that \texttt{XSum} contains 53\% of the tokens from our dataset, while \texttt{XL-Sum} contains 58\% of the tokens from our dataset across all languages present in \texttt{M3LS}. However, they are uni-modal in nature, while \texttt{XSum} is uni-lingual. 
We observe that \texttt{M3LS} is magnitudes larger when compared to \texttt{XSum}, while exceeding by times 2-3 in almost all individual language instances when compared to \texttt{XL-Sum}. Both of these datasets are used to train and fine-tune several state-of-the-art-summarization models like \texttt{Pegasus}, hence we believe that \texttt{M3LS} will offer a wider and better language modelling support in terms of size and diversity for the languages present in it, with the additional benefit of multi-modality.

    

\section{Experiments}
\subsection{Setup}
Depending upon the number of instances in each language within \texttt{M3LS}, we perform a train:test:validation split with a ratio of 80:10:10 if the number of instances is below 50K and 90:5:5 otherwise. To conduct our experiments in a multi-lingual setting, we survey publicly available tokenizers and sentence segmenters for multiple languages, and we combine them within one dedicated package for our experiments. We further define a set of rules for sentence segmentation for languages lacking such support from external packages within our package\footnote{\url{https://github.com/zenquiorra/TokSeg}}.

We compile our package using \texttt{segtok}\footnote{\url{https://pypi.org/project/segtok/1.1.0/}} for the Indo-European language, \texttt{IndicNLP}\footnote{\url{https://github.com/anoopkunchukuttan/indic_nlp_library}} for Indian languages, \texttt{fugashi} \cite{mccann-2020-fugashi} for Japanese (\texttt{ja}) and \texttt{chinese}\footnote{\url{https://pypi.org/project/chinese/}} for Chinese. 

For data pre-processing steps such as stopword removal, we collect stopwords from the \texttt{nltk}\footnote{\url{https://nltk.org/}} package, and publicly available stopwords present in the \texttt{spaCy}\footnote{\url{https://github.com/explosion/spaCy/tree/master/spacy/lang}} repository for all languages in a centralized pipeline for our experiments.

We evaluated the performance of various summarization techniques utilizing our dataset, including simpler techniques such as \texttt{LEAD-3} and \texttt{RANDOM} which have proven to be quite useful in past \cite{ghalandari2020large,scialom2020mlsum,sharma2019bigpatent}. We have also included statistics based \texttt{CENTROID} \cite{radev2004centroid} and graph based \texttt{TextRank} \cite{mihalcea2004textrank} techniques. 

To have a fair comparison across multiple languages using a shared dedicated model, we have evaluated the performance of an abstractive technique in a multi-lingual setting utilizing a pre-trained checkpoint\footnote{\url{https://huggingface.co/csebuetnlp/mT5_multilingual_XLSum}} for summarization of the transformer-based  \texttt{MT5} \cite{xue2020mt5} model. Finally to explore the multi-modal aspect of our dataset, we evaluate the performance of a multi-modal encoder-decoder based technique \cite{zhu2018msmo} that utilizes images and text to generate a multi-modal text summary. However the publicly available implementation\footnote{We use the implementation provided by the authors, which is a multi-layered package, modification of which to be compatible for a multi-lingual setting isn't feasible based on the software complexity} for \texttt{MSMO} restricts us to evaluate it only for the English language. However, to compare this score, we evaluate the performance of three state-of-the-art transformer-based models - \texttt{Pegasus} \cite{zhang2020pegasus}, \texttt{BART} \cite{lewis2020bart}, and \texttt{T5} \cite{xue2020mt5} for summarization which are compatible with the English language. 

Since, two of the pre-trained models we described above are fine-tuned on \texttt{XSum} and \texttt{XL-Sum} datasets which are extracted from the same source - \texttt{BBC News} - we avoid fine-tuning on models to have a fair comparison of the models and we explain the scores in discussions.

In all techniques, we set the generated summary length threshold as the average length of gold summary for the corresponding language in our corpus.

\subsection{Baselines}
\textsc{\bf Simpler Extractive Approaches}

\textsc{LEAD-3}: In this baseline, the first three sentences of the source text are extracted as the final summary. This method is a robust baseline, as shown by \cite{sharma2019bigpatent} for news summarization datasets. 

\textsc{RANDOM}: We recursively extract words randomly from the source text until the threshold summary length is reached. The aim of this baseline is to understand and compare other baselines with an unbiased model as a point of reference.

\noindent\textsc{\bf Statistical Approach}

\textsc{CENTROID}:  We use the strategy proposed by \citet{radev2004centroid}, which ranks sentences based on the centrality scores obtained by the words in the sentence. We use TF-IDF scores to measure each word's similarity, and extract top sentences from each ranking until the threshold summary length is obtained.

\noindent\textsc{\bf Graph Based Approach}

\textsc{TextRank}:  TextRank \cite{mihalcea2004textrank} is an unsupervised graph-based ranking technique based on the relevance of sentences in the source text\footnote{We use the implementation provided by the \texttt{gensim}{\url{https://radimrehurek.com/gensim_3.8.3/summarization/summariser.html} package and modify the segmentation and tokenizer part using our dedicated package.}} We consider the sentences which are most central to the document based on the ranking as generated summaries. 

\noindent\textsc{\bf RNN Based Approach}

\textsc{MSMO}: MSMO \cite{zhu2018msmo} is an encoder-decoder model trained for multi-modal summarization. It utilizes a multi-modal attention mechanism to generate multi-modal summaries utilizing text and images.

\noindent\textsc{\bf Transformer Based Approaches}

\textsc{MT5}: MT5 \cite{xue2020mt5} is a transformer-based seq2seq model pre\-trained for multiple natural language tasks. We use the publicly available checkpoint\footnote{\url{https://huggingface.co/csebuetnlp/mT5_multilingual_XLSum}} pre-trained for text summarization on the \texttt{XL-Sum} dataset \cite{hasan2021xl} for a multi-lingual setting.


\textsc{PEGASUS}: Pegasus \cite{zhang2020pegasus} is a transformer-based model, pre-trained on a task to remove  meaningful sentences from an input text, making it suitable for summarization. We used a checkpoint\footnote{\url{https://huggingface.co/google/pegasus-xsum}} of \texttt{PEGASUS} model pre-trained on the XSum dataset \cite{narayan2018don} for summarization.

\textsc{BART}: BART \cite{lewis2020bart} uses a standard seq2seq architecture with a bi-directional encoder and a left-to-right decoder. We use a pre-trained model trained on the DailyMail/CNN \cite{Nallapati2016AbstractiveTS} for our evaluation.

\textsc{T5}: T5 \cite{raffel2020exploring} is an encoder-decoder model trained on a mixture of natural language tasks, including translation and summarization; it converts any task into a text-to-text format. We use the pre-trained \texttt{T5-large} model for the summarization task.

\begin{table}[]

\caption{Comparison of ``ROUGE" f\-scores for summaries generated using Multi-modal baseline \texttt{MSMO} and Uni\-modal transformer based baselines against gold summaries from the English language dataset . ``R-f1" denotes ROUGE-1 f\-score, ``R-f2" denotes ROUGE\-2 f\-score, ``R-fL" denotes the ROUGE\-L f\-score, and ``BrS" denotes BERTSCORE.}
\centering
\begin{tabular}{|l|c|c|c|c|}
    \hline
    \textbf{English} & \textbf{R-f1}  & \textbf{R-f2} & \textbf{R-fL} & \textbf{BrS} \\
      \hline
      \texttt{BART} & 0.195 &  0.031 &  0.131 & 0.863 \\
      \texttt{Pegasus} & \textbf{0.389} &  \textbf{0.181} &  \textbf{0.321}  & \textbf{0.910} \\
      \texttt{T5}  & 0.197   &  0.0328 &  0.131 & 0.858 \\
      \texttt{MSMO} & 0.217 &  0.046 &  0.158 & 0.851 \\
    \hline
\end{tabular}
\label{tab:english-results} 
\end{table}

\section{Results and Discussion}

We evaluate the generated summaries against the gold summaries using the \texttt{ROUGE} \cite{lin2004rouge} evaluation metric. We report the \texttt{ROUGE-1}, \texttt{ROUGE-2}, \texttt{ROUGE-L} f-scores across every baseline discussed above \cite{lin2004rouge} (refer to Tables \ref{tab:comparison} and \ref{tab:tab-scores}). We additionally report \texttt{BERTSCORE} for English baselines \cite{zhang2019bertscore} (refer to Table \ref{tab:comparison}).


\begin{table*}[]

\caption{Performance of various techniques for summarization against the \texttt{M3LS} dataset gold summaries for every language. ``Lang" refers to the language code for a language according to the ISO 639-1 standard, ``R-f1" refers to the \texttt{ROUGE-1} f-scores, ``R-f2" refers to the \texttt{ROUGE-2} f-scores, ``R-fL" refers to the \texttt{ROUGE-L} f-scores}
\resizebox{\textwidth}{!}{
\begin{tabular}{lrrrrrrrrrrrrrrr}
\toprule
Base & \multicolumn{3}{c}{\textbf{Random}} & \multicolumn{3}{c}{\textbf{LEAD-3}} & \multicolumn{3}{c}{\textbf{TextRank}} & \multicolumn{3}{c}{\textbf{CENTROID}} & \multicolumn{3}{c}{\textbf{MT5}} \\
Lang  &  R-f1 &        R-f2 &        R-fL &        R-f1 &        R-f2 &        R-fL &        R-f1 &        R-f2 &        R-fL &        R-f1 &        R-f2 &        R-fL &  R-f1 &        R-f2 &        R-fL  \\
\midrule

bn &  0.003 &  0.000 &  0.002 &  0.001 &  0.000 &  0.000 &    0.000 &  0.000 &  0.000 &    0.000 &  0.000 &  0.000 & \textbf{ 0.004} &  0.001 &  0.003 \\
mr &  0.013 &  0.000 &  0.012 &  0.041 &  0.005 &  0.040 &    0.025 &  0.002 &  0.025 &    0.006 &  0.001 &  0.006 & \textbf{ 0.044} &  0.005 &  \textbf{0.044} \\
gu &  0.014 &  0.001 &  0.014 & \textbf{ 0.039} &  0.005 &  0.038 &    0.014 &  0.001 &  0.014 &    0.016 &  0.002 &  0.016 &  0.036 &  0.005 &  0.036 \\
ps &  0.002 &  0.000 &  0.001 &  0.000 &  0.000 &  0.000 &    0.002 &  0.000 &  0.001 &    0.000 &  0.000 &  0.000 & \textbf{ 0.003} &  0.000 &  0.001 \\
uk &  0.030 &  0.002 &  0.029 &  0.062 &  0.016 &  0.061 &    0.043 &  0.010 &  0.042 &    0.032 &  0.006 &  0.032 &  \textbf{0.094} &  0.025 & \textbf{ 0.094} \\
pt &  0.179 &  0.009 &  0.114 &  0.204 &  0.033 &  0.124 &  0.199 &  0.030 &  0.128 &    0.089 &  0.008 &  0.075 &  \textbf{0.276} &  0.085 &  0.193 \\
id &  0.118 &  0.001 &  0.083 &  0.172 &  0.037 &  0.117 &    0.144 &  0.030 &  0.104 &    0.104 &  0.014 &  0.080 & \textbf{ 0.289 }&  0.115 &  0.233 \\
ne &  0.000 &  0.000 &  0.000 &  0.000 &  0.000 &  0.000 &    0.000 &  0.000 &  0.000 &    0.000 &  0.000 &  0.000 &  0.000 &  0.000 &  0.000 \\
pa &  0.012 &  0.000 &  0.012 &  \textbf{0.038} &  0.004 & \textbf{0.038} &    0.014 &  0.001 &  0.014 &    0.010 &  0.002 &  0.010 &  0.026 &  0.000 &  0.026 \\
si &  0.014 &  0.000 &  0.014 &  0.032 &  0.004 &  0.031 &    0.019 &  0.002 &  0.019 &    0.007 &  0.001 &  0.007 &  \textbf{0.039} &  0.018 &  \textbf{0.039} \\
ur &  0.006 &  0.000 &  0.006 &  0.023 &  0.001 &  0.023 &    0.006 &  0.000 &  0.005 &    0.024 &  0.001 &  0.023 &  \textbf{0.044 }&  0.000 &  \textbf{0.044} \\
fr &  0.168 &  0.007 &  0.107 &  0.206 &  0.043 &  0.126 &    0.177 &  0.033 &  0.115 &    0.164 &  0.024 &  0.110 &  \textbf{0.209} &  0.041 &  0.141 \\
ru &  0.032 &  0.001 &  0.032 &  0.071 &  0.017 &  0.069 &    0.041 &  0.012 &  0.040 &    0.036 &  0.008 &  0.036 &  \textbf{0.081 }&  0.011 &  \textbf{0.081} \\
ja &  0.069 &  0.001 &  0.068 &  0.126 &  0.012 &  0.120 &    0.084 &  0.007 &  0.081 &    0.063 &  0.004 &  0.062 &  \textbf{0.306} &  0.081 &  0.291 \\
te &  0.010 &  0.000 &  0.009 &  0.023 &  0.001 &  0.023 &    0.011 &  0.000 &  0.011 &    0.008 &  0.001 &  0.008 &  \textbf{0.026} &  0.000 &  \textbf{0.026} \\
ta &  0.014 &  0.001 &  0.014 &  \textbf{0.034 }&  0.005 &  \textbf{0.034} &    0.023 &  0.003 &  0.022 &    0.012 &  0.001 &  0.012 &  0.026 &  0.000 &  0.026 \\
zh &  0.022 &  0.001 &  0.022 &  \textbf{0.053} &  0.008 &  0.051 &    0.042 &  0.005 &  0.041 &    0.025 &  0.003 &  0.025 &  0.125 &  0.042 &  0.118 \\
es &  0.177 &  0.008 &  0.117 &  0.180 &  0.033 &  0.117 &    0.110 &  0.018 &  0.073 &    0.081 &  0.008 &  0.067 &  \textbf{0.280} &  0.084 &  0.202 \\
hi &  0.010 &  0.000 &  0.010 &  \textbf{0.018} &  0.002 &  0.018 &    0.013 &  0.001 &  0.013 &    0.005 &  0.000 &  0.005 &  0.002 &  0.000 &  0.001 \\
en &  0.146 &  0.002 &  0.102 &  \textbf{0.175} &  0.026 &  0.114 &    0.100 &  0.014 &  0.071 &    0.140 &  0.016 &  0.102 &  \textbf{0.427} &  0.182 &  0.345 \\

\bottomrule
\end{tabular}
}
\label{tab:tab-scores}
\end{table*}

\subsection{Multi-lingual baseline scores}
We observe that transformer based techniques used in our experiments perform significantly better compared to other techniques. However, for the ``MT5" column, we observe very high scores and spikes of very low scores as observed in Table \ref{tab:tab-scores}, this behavior maybe caused due to two factors:

\begin{itemize}
    \item Relatively high scores can be attributed to the use of ``MT5" checkpoint that is fine-tuned for the task of summarization on a dataset (\texttt{XL-Sum}) obtained from same source as ours. 
    \item Very low scores for some languages can be attributed to the ``ROUGE" evaluation metric which relies on token overlap\footnote{We are not implementing stemming of tokens during evaluation, due to the lack of support of multi-lingual stemming methods across various softwares which we have used for experimentation and to have an even comparison with the supported languages}. Many of these languages, especially the ones with \texttt{Dravidan} and \texttt{Indo-European} origins have words which change their form significantly depending on their placement in the text and the context in which they appear, hence simple token overlap metrics show lower scores if the root form of the word isn't considered.
\end{itemize}

We observe that \texttt{LEAD-3} performs better for the languages in which transformer-based baseline performs poorly, this can be attributed to two factors:
\begin{itemize}
    \item As shown by \citet{sharma2019bigpatent} that \texttt{LEAD-3} performs very well for summarization tasks when we consider the news domain, suggesting the idea that top sentences capture a lot of information within a news article.
    \item \texttt{LEAD-3} considers top-3 sentences from the text, unlike abstractive summarization, new tokens or new forms of existing tokens are not present in the given article. Since it is an extractive technique, the chances of token overlap are higher and hence better ``f-scores".
\end{itemize}

\subsection{Multi-modal baseline scores}

Due to the limitation of lack of pre-trained frameworks in a multi-modal setting for most of the languages in the dataset, we were constrained to evaluate the multi-modal technique on the \texttt{English} dataset. On comparing the ``f-scores" of various uni-modal techniques with the multi-modal technique, we notice that the transformer based model \texttt{Pegasus} outperforms other techniques. This is largely attributed to the fact that the pre-trained checkpoint we have used for evaluation of summaries through the \texttt{Pegasus} model is fine-tuned on the \texttt{XSum} dataset, which has data collected from the same source as ours. We observe that for other models which are not fine-tuned on a dataset extracted from same source as ours, the multi-modal technique \texttt{MSMO} is able to outperform other techniques.
 
\subsection{Abstractiveness of the proposed dataset}

We propose an abstractive summarization dataset where the target summaries are manually written by human beings. The \texttt{M3LS} dataset demands abstractive techniques since the percentage of novel uni-grams in the dataset is quite high (refer to ``abs.gold" column in Appendix \ref{sec:appendix-B}). This fact is also observed in the results from the baseline techniques. For instance, \texttt{MT5} performs consistently superior for multiple languages as observed in Table\ref{tab:tab-scores}, the abstractive baselines have thrice as good ROUGE scores as the extractive baselines. 

\section{Conclusion}

In this work, we release a large-scale multi-modal multi-lingual summarization dataset comprising of over 1.1M+ news articles and spanning 20 languages, and motivate the problem statement of Multi-modal Multi-lingual summarization using \texttt{M3LS}. To the best of our knowledge, this is the first ever multi-modal summarization data set spanning several languages.  The proposed dataset is the largest summarization dataset for 13 out of  20 languages. We have evaluated the performance of various baselines to establish the quality of the proposed dataset in both  multi-modal and  multi-lingual settings. 
Through this work, we hope to instigate research in various less-explored languages in the community for various research problems including but not limited to summarization, headline generation, keyword extraction, image caption generation, multi-modal embedding generation, etc. In future works, we plan to work on shared models which address the \texttt{M3LS} task utilizing our dataset. 

\section*{Limitations}
There are a few considerations to keep in mind in our work. \textbf{First}, the dataset currently has a multi-modal input, mapping to a textual summary. However, future work could involve annotating images to enhance the dataset with a multi-modal output. \textbf{Second}, the distribution for languages in the \texttt{M3LS} dataset is skewed due to the imbalanced number of articles published in BBC across languages and the late establishment of virtual print media in certain languages (as shown in Appendix \ref{sec:appendix-A}). \textbf{Third}, the current dataset uses an independent identically distributed split to create train and test sets, but more advanced techniques such as adversarial splits and likelihood splits could also be explored in future work. \textbf{Fourth}, while the current manuscript does not evaluate the dataset on both multi-modal and multi-lingual aspects simultaneously, we believe that this dataset has the potential to contribute to the development of such systems in the future.

\section*{Acknowledgements}
This publication is an outcome of the R\&D work undertaken in the project under the Visvesvaraya Ph.D. Scheme of Ministry of Electronics \& Information Technology, Government of India, being implemented by Digital India Corporation (Formerly Media Lab Asia).

\bibliography{anthology,main}
\bibliographystyle{acl_natbib}

\newpage

\appendix

\section{Frequency of number of articles present in the \texttt{M3LS} dataset for a given year}
\label{sec:appendix-A}

\begin{figure}[h]
    \centering
    \includegraphics[width=0.5\textwidth]{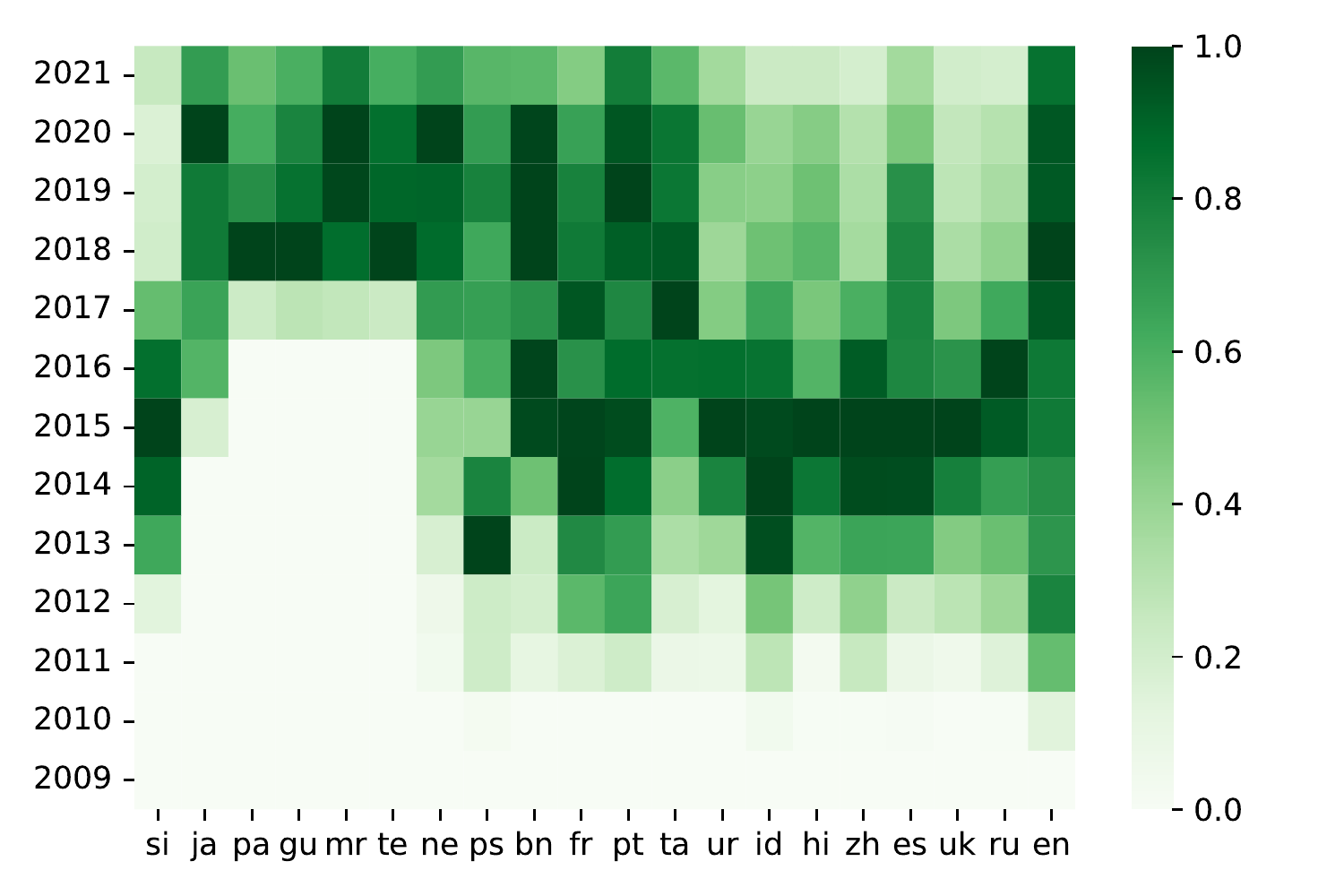}
    \newline
    \caption{Temporal span of each language in the \texttt{M3LS} dataset. Darker colors correspond to a score of 1.0 and vice versa, which indicates a higher frequency of the number of articles published during the year for that language.}
    \label{fig:temporal-plot}
\end{figure}

\section{Detailed Statistics of \texttt{M3LS} dataset} \label{sec:appendix-B}

The detailed statistics of the curated \texttt{M3LS} dataset are presented in Table \ref{tab:detailed-stats}. 

\begin{table*}[]
\centering
\caption{``Lang." represents the language code from the IS0 639-1 standard, \textbf{articles} represents the number of articles for every language in the corpus, \textbf{toks.} represents the average number of tokens in an article within the corpus for the language, \textbf{tok.uni.} represents the average number of unique tokens in any article for the language, \textbf{sum.tok.} represents the average number of tokens in summary of an article for the language, \textbf{sum.uni} represents the average number of unique tokens within the summary of an article for the language, \textbf{sent.} represents the average number of sentences in a given article for the language, \textbf{abs.gold} represents the average percent of tokens in summary which are absent from the article for a given language, \textbf{images} represents the average number of images in an article for the given language, \textbf{i.c.r} represents the average ratio of the number of images consisting of a caption attached to them against the number of images which do not have a caption with them.}
\label{tab:detailed-stats}
\resizebox{\textwidth}{!}{
\begin{tabular}{lrrrrrrrrr}
\toprule
\textbf{Lang.} &  \textbf{articles} &   \textbf{tok.} &   \textbf{tok.uni.} &  \textbf{sum.tok.} &  \textbf{sum.uni.} &  \textbf{sent.} &  \textbf{abs.gold} &  \textbf{images} &  \textbf{i.c.r} \\
\midrule
bn &     25283 &   464.78 &  254.73 &           23.12 &          21.97 &      28.42 &     43.18 &    2.44 &             0.58 \\
mr &     16161 &   871.36 &  404.82 &           27.88 &          25.02 &      63.32 &     48.33 &    0.00 &             0.00 \\
gu &     12175 &   868.75 &  404.84 &           25.87 &          23.63 &      50.11 &     40.87 &    4.94 &             0.30 \\
ps &     23205 &   523.35 &  207.95 &           32.85 &          27.52 &      18.29 &     33.32 &    2.10 &             0.49 \\
uk &     90846 &   471.90 &  216.57 &           24.59 &          22.78 &      23.49 &     54.07 &    1.56 &             0.22 \\
pt &     39454 &  2855.68 &  424.14 &           38.57 &          33.01 &     114.22 &     35.23 &    3.34 &             0.64 \\
id &     56108 &   587.88 &  225.12 &           24.59 &          22.89 &      28.54 &     37.07 &    2.47 &             0.57 \\
ne &     18953 &   402.17 &  229.71 &           21.50 &          20.95 &      23.79 &     45.00 &    2.07 &             0.24 \\
pa &     11600 &   843.74 &  319.75 &           30.96 &          27.35 &      38.87 &     30.42 &    4.57 &             0.37 \\
si &     10148 &   331.55 &  186.26 &           24.30 &          23.43 &      15.53 &     51.33 &    1.67 &             0.34 \\
ur &     55107 &   690.47 &  264.57 &           35.95 &          31.00 &       1.07 &     27.94 &    2.35 &             0.43 \\
fr &     25923 &   413.45 &  179.22 &           31.22 &          27.26 &      14.43 &     40.75 &    1.64 &             0.47 \\
ru &     95345 &   668.61 &  302.63 &           28.38 &          25.74 &      26.68 &     52.80 &    1.96 &             0.38 \\
ja &     11023 &  1052.94 &  282.58 &           48.59 &          36.97 &      33.92 &     29.86 &    2.90 &             0.56 \\
te &     15511 &   626.24 &  353.70 &           23.71 &          21.39 &      51.72 &     52.19 &    4.41 &             0.27 \\
ta &     38523 &   354.35 &  209.30 &           21.18 &          19.91 &      23.39 &     56.34 &    2.55 &             0.27 \\
zh &     60830 &   787.68 &  309.11 &           34.38 &          29.80 &      29.87 &     37.21 &    2.00 &             0.48 \\
es &     66816 &  2649.57 &  345.35 &           30.64 &          26.28 &      83.22 &     38.43 &    4.15 &             0.66 \\
hi &     61852 &   776.39 &  265.10 &           29.45 &          25.70 &      37.21 &     32.00 &    1.24 &             0.0 \\
en &    376367 &   657.95 &  268.60 &           25.27 &          23.40 &      22.28 &     32.86 &    1.39 &             0.35 \\
\bottomrule
\end{tabular}
}
\label{tab:data-stats}
\end{table*}

\end{document}